\documentclass{article}

\usepackage{arxiv}

\usepackage[utf8]{inputenc} 
\usepackage[T1]{fontenc}    
\usepackage{hyperref}       
\usepackage{url}            
\usepackage{booktabs}       
\usepackage{amsfonts,amsmath,amssymb}       
\usepackage{nicefrac}       
\usepackage{microtype}      
\usepackage{graphicx}
\usepackage{algorithmic}
\usepackage[ruled,linesnumbered]{algorithm2e}
\usepackage[dvipsnames]{xcolor}

\newcommand{\ie}{\textit{i.e.,}}
\newcommand{\eg}{\textit{e.g.,}}
\newcommand{\commenttxt}[1]{}

\renewcommand{\mytagline}{} 

\newtheorem{definition}{Definition}

\title{Evolution of Data-driven Single- and Multi-Hazard Susceptibility Mapping and Emergence of Deep Learning Methods}
\author{Jaya~Sreevalsan-Nair and Aswathi~Mundayatt\thanks{Corresponding author: \texttt{\{jnair,aswathimundayatt.valappil003\}@iiitb.ac.in} 
  \\
  Graphics-Visualization-Computing Lab,\\
  International Institute of Information Technology Bangalore, Karnataka 560100, India \\
  \texttt{http://www.iiitb.ac.in/gvcl} 
}}

\begin{document}
\maketitle

\begin{abstract}
  Data-driven susceptibility mapping of natural hazards has harnessed the advances in classification methods used on heterogeneous sources represented as raster images. Susceptibility mapping is an important step towards risk assessment for any natural hazard. Increasingly, multiple hazards co-occur spatially, temporally, or both, which calls for an in-depth study on multi-hazard susceptibility mapping. In recent years, single-hazard susceptibility mapping algorithms have become well-established and have been extended to multi-hazard susceptibility mapping. Deep learning is also emerging as a promising method for single-hazard susceptibility mapping. Here, we discuss the evolution of methods for a single hazard, their extensions to multi-hazard maps as a late fusion of decisions, and the use of deep learning methods in susceptibility mapping. We finally propose a vision for adapting data fusion strategies in multimodal deep learning to multi-hazard susceptibility mapping. From the background study of susceptibility methods, we demonstrate that deep learning models are promising, untapped methods for multi-hazard susceptibility mapping. Data fusion strategies provide a larger space of deep learning models applicable to multi-hazard susceptibility mapping.
\end{abstract}


\section{Introduction}\label{sec:introduction}

Natural hazards or disasters are one of the leading causes of concern for developing countries despite the reduction in mortality owing to advancements in early prediction, resilient infrastructures, and effective response systems~\cite{ritchie2022natural}. It is still debatable if there is an increasing trend in the frequency of natural disasters~\cite{alimonti2023number} and if climate change influences the frequency~\cite{wen2023disaster}. Notwithstanding the debate, there is a continued need to improve the study of natural disasters to reduce the drastic effects of natural disasters, \ie injuries, homelessness, habitat displacements, economic losses, etc. Nowadays there is increased awareness of any geographical region being susceptible to multiple hazards~\cite{stalhandske2024global}. This has been made possible by the use of various prediction models. One such prediction model leads to \textit{multi-hazard susceptibility mapping}. There are tremendous advancements in the susceptibility mapping of single events, such as floods, landslides, forest fires, etc., which can now be extended to the computation of multi-hazard susceptibility maps (MHSMs).

\begin{definition} A \textbf{\textit{hazard susceptibility map}} is a mathematical function that provides the probability and severity of the occurrence of a natural hazard at a geographical location based on the spatial distribution of causative (or conditioning) factors and instances of historical occurrences of the hazard in a region containing the location.
\end{definition}

Given that the hazard susceptibility map gives a numerical or categorical value for probability or severity, respectively, at geographical locations, these maps are best visualized as geographical maps or cartographs.

Historically, the initial focus of the multi-hazard analysis was on determining multi-layer single-hazard risk assessment which combines risks from multiple single hazards~\cite{stalhandske2024global}. Since it did not account for temporal sequence or co-occurrence of events, the studies have increasingly shifted to multi-hazard risk assessment which accounts for the interactions of events of multiple single hazards. For instance, compound flooding occurs when the same region is susceptible to different kinds of flooding~\cite{hendry2021compound}, \ie coastal, pluvial, fluvial, and flash floods~\cite{dewan2013floods}. The Intergovernmental Panel on Climate Change (IPCC) defines \textit{compounding events} to be events that occur simultaneously or successively, a combination of events with compounding effects on impact, or a combination of events that cause an extreme event in its wake.

Here, we focus on the use of data science methods for MHSM computation. Data science provides appropriate tools that can use the causative and historical instances as inputs to provide the susceptibility value as output. This paper focuses on the evolution of data science methods for a single-hazard susceptibility mapping determination and its adaptation for MHSM computation. We elaborate on the evolution of these methods for floods and landslides since they are widely studied.

We observe from the research literature that the data-driven methodology of susceptibility mapping focuses on a large proportion of case studies pertaining to local and regional study areas, as opposed to national~\cite{bentivoglio2022deep}. This can be considered to be an organic choice of studies, given the \textit{spatial heterogeneity} of (co-)occurrences of natural hazards~\cite{yang2024novel}. Since our work is not intended to be a systematic literature review, this paper will not provide an exhaustive coverage of all case studies. The focus of our work is on studying the patterns in popularity of methodologies for a single hazard, for which we will constrict to a few case studies as examples.

To the best of our knowledge, a study on the evolution of methods across multiple single hazards (such as floods and landslides) has not been jointly reported. As a novelty, this paper also steers their extensions to multi-hazard susceptibility mapping and discusses the overall emergence of deep learning models in the context of evolution and extension. Our contributions are:
\begin{itemize}
  \item to report the evolution of the methods used in single-hazard susceptibility mapping with a focus on floods and landslides, 
  \item to discuss the state-of-the-art methods used in multi-hazard susceptibility mapping as an extension of single-hazard susceptibility mapping, 
  \item to discuss the emergence of deep learning methods for both single and multi-hazard susceptibility mapping.
\end{itemize}

\section{Background on Single-Hazard Susceptibility Mapping}\label{sec:shsm}
The methodology for MHSM determination is derived from the widely used one for single hazard susceptibility mapping, such as that for floods, landslides, forest fires, etc.~\cite{pourghasemi2020assessing}. The widely used flood susceptibility mapping (FSM) and landslide susceptibility mapping (LSM) provide appropriate spatial distribution models, which are also used for practical purposes. Hence, we study such single-hazard susceptibility mapping methods, namely, statistical analysis (SA), multi-criteria decision analysis (MCDA), soft computing (SC), metaheuristic optimization (MHO), machine learning (ML), and deep learning (DL) models. Here, we focus on ML/DL models owing to their increasing usage for susceptibility mapping. We first take the examples of FSM and LSM computation~\cite{tay2023disaster} for a comparative analysis leading to a high-level generalization.

A comparison of FSMs and LSMs demonstrates their similarities, thus providing a high-level generalized algorithm for a single-hazard susceptibility mapping (Algorithm~\ref{alg:shsm}). This algorithm is distilled from a systematic review of the state-of-the-art and comparative analyses for FSM~\cite{bentivoglio2022deep,khosravi2019comparative} and LSM computing~\cite{fang2021comparative,fleuchaus2021retrospective}. We observe that the same procedure is even followed for a few of the relatively rarer hazards, \eg bushfires~\cite{tehrany2021application}.

In data-driven methods, the labeled data and model validation are critical. The former provides the patterns for training or building a model in the algorithm, and the latter assesses the credibility of the model. For susceptibility mapping, the labels of occurrence and non-occurrence of hazards are needed for training the models to ensure class balance. The occurrences are determined using past observations or measurements from remote sensing and local sensing stations, and the non-occurrence labels are obtained by randomly picking locations that never had a history of the hazard occurrence~\cite{bentivoglio2022deep}. Many of these models are then validated using AUC (area under the curve) metrics for the Receiving Operating Characteristic (ROC) curve, Success Rate Curves (SRC), or Prediction Rate Curves (PRC)~\cite{fleuchaus2021retrospective}. Generally for all single hazards, despite the popularity of data-driven susceptibility mapping, there are no standardizations in the process or the outcome~\cite{fleuchaus2021retrospective}.

\begin{algorithm}[t]
  \small
  \caption{\label{alg:shsm} A High-level Generalized Algorithm for Single-Hazard Susceptibility Mapping~\cite{tay2023disaster}}
  \DontPrintSemicolon
  \SetAlgoLined
  \SetKwInOut{Input}{Input}
  \SetKwInOut{Output}{Output}
  \Input{A stack of raster images of the map of the study area corresponding to concerned conditioning (or causative) factors}
  \Input{A set of historical sites, $S_H$ of occurrence as well as non-occurrence of the hazard, with sufficient class balance}
  \Output{A raster image of the map of the study area, $O_r$, colored based on the probability of occurrence of hazard}
  \Output{Validation metrics of the susceptibility map}
  Split $S_H$ to training $S_{tr}$ and testing $S_{tt}$ sets (\eg 70/30 or 80/20 split)\;
  Select features using multi-collinearity test or similar tests\;
  Run a computational model (SA, MCMDA, SC, MHO, ML, or DL) or an ensemble of such models using $S_{tr}$ and input raster images of selected features to classify pixels in the output raster image, $O_r$\;
  \If {an ensemble of models}
      {
        Run a meta-model to combine the results of the ensemble
        }
  Run one or more validation methods using $S_{tt}$ to compute metrics (\eg AUC-ROC curve, AUC-SRC, or AUC-PRC)\;
  \KwRet $O_r$\;
  \end{algorithm}

\subsection{Flood Susceptibility Maps}\label{sec:fsm}
Floods are of different types, namely, river, flash, coastal, dam breaks, pluvial, and urban~\cite{dewan2013floods}. Pluvial floods may be considered urban floods~\cite{bentivoglio2022deep} for FSMs. For all flood types, the hazard maps are studied using simulations, and the inundation and susceptibility maps are analyzed based on observations~\cite{bentivoglio2022deep}. These map-type-based methodologies are also spatial-scale-agnostic, where scales refer to local, regional, national, and supra-national. However, there are limited studies for the national-level and no studies at all for the supra-national-level (\ie continental-level) FSMs. Thus, computing FSMs at larger spatial scales provides scope for future work.

In the initial work on FSM computation, numerical nonlinear models were used for watersheds. However, these physically based hydrological models increasingly are unable to capture the complexity of the watersheds~\cite{khosravi2019comparative}. Hence, data-driven models began to be increasingly used for several case studies. In roughly chronological order of evolution, these models are bivariate statistical, multivariate statistical, soft computing, multi-criteria decision-making, and machine learning models. We redirect the readers to a more detailed explanation of these methods in the systematic literature review by Kaya and Derin~\cite{kaya2023parameters}. A slice of the review reflecting the evolution is given below:
\begin{itemize}
\item The hydrological models include HEC-RAS, which is popularly used and continues to be used for large-scale areas in developing countries.
\item Frequency ratio (FR) and logistic regression (LR) are popularly used bivariate statistical analysis (BSA) and multivariate analysis (MSA) used for FSM computation. The statistical methods largely captured only the linear characteristic of FSMs through the correlation between conditioning factors and the floods. However, the characterization of the nonlinear aspect of the complexity of flooding is ignored in these models.
\item Multi-criteria decision-making (MCDM) is a process where multiple criteria are jointly evaluated. It provides a more realistic modeling of complex phenomena, such as disasters. MCDM is implemented using MCDA models which are effective for FSM computation through effective prioritization of alternative decisions. Analytic Hierarchy Process (AHP) is a widely used MCDA technique for FSMs. MCDM methods are simple and continue to be relevant, however, these methods require an expert opinion, making them highly subjective.
\item Soft computing methods can characterize the fuzziness in real-life events, such as floods. These methods include fuzzy logic, genetic algorithms, and neural networks, which complement each other and are used in combination. The use of artificial neural networks (ANNs) also paves the way for the recent spurt in the use of ML/DL models.
\item A class of methods for nonlinear modeling, namely, metaheuristic algorithms, is effectively used for FSM. Metaheuristic optimization algorithms, such as particle swarm optimization (PSO), gray wolf optimization (GWO), and bat algorithm, are used for FSM.
\item Current methods use machine learning, deep learning, and big data frameworks, thus shifting away from traditional expert-based systems. Random forest and convolutional neural networks (CNNs) are currently popularly used for FSMs. These also include the use of ensemble methods~\cite{costache2020novel}. ML models require appropriate feature engineering, which is avoided in DL models~\cite{bentivoglio2022deep}. However, DL models are more data- and resource-hungry to automatically detect the embeddings present in the data, and their inter-relationships. There are several comparative analyses in recent literature, where in some cases, the ML models are superior to others~\cite{nachappa2020flood}, and not so in other cases~\cite{pham2021comparison}.
\item Finally, hybrid models, that use an integration of the aforementioned individual models, improve the prediction accuracy of FSMs. As an example of a hybrid model across different classes of methods, deep learning ensemble models with AHP and FR are effective for FSM computation for flash floods~\cite{costache2020novel}. Ensembles could also be built within the same class of methods, \eg BSA and MSA are combined through the use of FR and LR for effective FSM computation~\cite{tehrany2014flood}.
\end{itemize}

While the work on the choice of models has stabilized over time, there are variations in the choice of the conditioning factors (also called as causative factors or parameters), as found in a recent meta-analysis of 170 studies~\cite{kaya2023parameters}. The factors fall under different classes, namely, environmental, hydrological, topographic, and sociodemographic. The disparity in the usage of conditioning factors could be due to three reasons. Firstly, these factors depend on the physical and natural characteristics of the locality of the study area. Secondly, the data availability influences the adoption of these factors in each study area. Thirdly, the susceptibility mapping is not always intended to be holistic, \eg susceptibility maps based on hydrological processes can be combined with vulnerability maps. That said, all studies in the meta-study used at least one conditioning factor from each of the different classes.

An unintended consequence of this variety in the usage of conditioning factors is that the significant or effective factors vary across the selected studies. This is a critical point, as the significance of the factors determines the outcome of the data-driven methodology, and \textit{precedence} is often used for designing susceptibility mapping for both single and multiple hazards.

\subsection{Landslide Susceptibility Maps}\label{sec:lsm}
We observe that the evolution of methods in FSM computation is the same for LSM~\cite{fang2021comparative}, except that the hydrological models are replaced by deterministic methods. Deterministic methods, similar to hydrological models, apply physical laws accurately to determine slope instabilities. But these methods, again like the hydrological models, tend to be static and inflexible to characterize the complexity of the landslide process.

A recent study calls for the use of retrospective evaluation of landslide susceptibility maps based on recent events~\cite{fleuchaus2021retrospective}. This study further states that the reliance on AUC-based validations does not suffice owing to the high uncertainty in statistical analysis and local characteristics concerning geomorphology and landslide process. Hence, additional model validation including the local characteristics should be performed to improve the confidence of the LSM. Epistemic uncertainties in landslide susceptibility mapping arise from errors in reference landslide inventory errors, difficulties in causality analysis, and the choice of parameters for the LSM model~\cite{zezere2017mapping}. The retrospective evaluation and uncertainties are generalizable to other hazards, such as floods, etc.

Different from FSM, a few of the LSM models also use terrain mapping units to accurately characterize a surface deformation event, such as landslides~\cite{zezere2017mapping}. The type and size of the terrain mapping units influence the LSM results. 

In summary, we observe that LSM and FSM computations are predominantly similar with very few hazard-characteristic differences. Hence, we can conclude that the generalized algorithm (Algorithm~\ref{alg:shsm}) is applicable across several natural hazards at a high level.

\section{Background on Multi-Hazard Susceptibility Mapping}\label{sec:mhsm}

Using data-driven methodology, the MHSMs are predominantly derived from multiple single-hazard susceptibility maps computed using Algorithm~\ref{alg:shsm}. However, for some natural hazards, the susceptibility maps are equivalent to maps of their physical phenomena, which are computed using hazard-specific methods. Such hazards include earthquakes and shoreline erosion, among others. Once we provide a holistic view of the computation of single-hazard maps to be used in an MHSM, we study a few examples of methods used for MHSM computation. These examples help us arrive at a generalized algorithm that represents the state-of-the-art for MHSM computation.

\subsection{Other Relevant Single-Hazard Maps}
Here, we show examples of hazard maps that are used in MHSM computed and are generated differently from Algorithm~\ref{alg:shsm}. These are earthquakes and shoreline erosion.

A map of \textit{peak ground acceleration} (PGA) is treated as the hazard map for earthquakes~\cite{bordbar2022multi,pouyan2021multi}. The PGA map can be computed using different classes of methods, namely, experimental-statistical, probabilistic, and deterministic methods. Of these classes, probabilistic seismic hazard analysis (PSHA) is widely used. After PSHA was proposed in 1968, it has increasingly become a combination of data-driven strategy and physical model. Seismic parameters, such as earthquake magnitude, return period, PGA, etc. are relevant for the hazard map computation. PGA itself captures the maximum ground motion experienced at each location if an earthquake is expected there. PSHA is designed to identify springs, determine seismicity parameters, and calculate PGA using the precedence of earthquake events and available seismic parameters. PGA maps are computed using an earthquake in a certain period ($T_E$ years) in the future with a specific probability ($p_E$) and the average return period ($T_R$ years). Unlike FSMs and LSMs which can also have probability values, the PGA map always has categorical values based on ordinal classes, such as ``Low,'' ``Moderate,'' and ``High.''

For shoreline erosion mapping, a physical model called the digital shoreline analysis system (DSAS) is used~\cite{prasad2023multi}. In this model, there is significant uncertainty in automatic shoreline delineation owing to the land-water transition being occupied by a water-saturated zone. To address this uncertainty, tasseled cap transformation (TCT) is used which uses the satellite data for more accurate shoreline detection. Once the shoreline extraction is completed, the \textit{end point ratio} (EPR) is estimated as the distance of shoreline/coastline shifting to the time taken. EPR is then used for generating the shoreline erosion map.  

\subsection{State-of-the-Art MHSM Computation Methods}
We observe that the state-of-the-art MHSM computation methods compute the individual single-hazard susceptibility maps separately, and then combine them using \textit{late fusion} or \textit{lazy fusion} strategy, as used in ML/DL parlance. Late fusion implies that the decisions of the individual susceptibility maps are determined separately and then combined using several methods, \eg averaging, sampling, decision using AHP, etc. We also observe that the data-driven MHSM computation by using single susceptibility maps is a relatively recent practice, starting circa 2019, to the best of our knowledge.

The MHSMs are visualized as a cartograph or geographical map using the following strategies for color schemes: (i) a set of different combinations of the natural hazards, (ii) a bivariate matrix of severity, in the case of two natural hazards, and (iii) a single joint ranking of all hazards. For the first strategy, we have a maximum of $2^{N}$ classes of combinations of $N$ hazards in the MHSM~\cite{pourghasemi2020assessing,pouyan2021multi,piao2022multi}. The categorical color palette is then assigned to relevant classes. For the second strategy, we have severity classes in a matrix with \textit{Hazard-1} row-wise and \textit{Hazard-2} column-wise, and each matrix element is a class corresponding to a combination of the 2 hazards with specific severities, \eg \textit{(Low susceptibility Flood, Low intensity Landslide), (Low susceptibility flood, Moderate susceptibility landslide), etc.}~\cite{nicu2022multi,elia2023assessing,lombardo2020spatial,nachappa2020multi}. Either a categorical color palette for the matrix or a bivariate gradient color palette may be used. For the third strategy, the multi-hazard susceptibility is mapped to a numerical value, which allows for a joint measure and a single gradient color palette~\cite{rehman2022multi,karakas2023hybrid}.

We observe that there are three strategies of MHSM computation methods based on the choice of model used for multiple single-hazard susceptibility mapping: (i) a single selected model used for all hazards (\textbf{Single-model}), (ii) a single optimal model selected from an ensemble of models for all hazards (\textbf{Single-model-from-ensemble}), and (iii) hazard-specific optimal model selected from an ensemble of models or prior knowledge (\textbf{Optimal-models}). We provide examples for each of these strategies below.
\begin{itemize}
\item \textit{Single-model strategy}: An assembled FR-AHP method is used for computing the FSM and LSM, and they are combined using a weighted sum for deriving an MHSM for mountainous regions~\cite{rehman2022multi}. A generalized additive model (GAM) is used for computing probability maps leading to susceptibility maps of thaw slumps and thermo-erosion gullies, where the susceptibility maps are combined using a bivariate color scheme to compute the final MHSM, as used for cryospheric hazards in the Arctic~\cite{nicu2022multi}. This procedure is used for retrogressive thaw slumps and active layer detachments in Alaska for cryospheric hazards~\cite{elia2023assessing}. A similar strategy was used earlier with landslide and gully erosion maps for determining threats in a cultural heritage site~\cite{lombardo2020spatial}. In a study of coastal regions, AHP is used for computing separate susceptibility maps for landslide, torrential flood, tsunami, and soil erosion, and finally, these maps are combined using a weighted sum, where the weights are computed using AHP~\cite{krassakis2023multi}. Random forest classifier is used for susceptibility maps for floods, landslides, and forest fires, after prioritizing conditioning factors for each using the Boruta algorithm, and the final MHSM is computed using a sum, for urban regions~\cite{pourghasemi2020assessing}. CNNs are used for computing separate susceptibility maps for flash floods, debris flows, and landslides, and then the 2-class maps are used to get combinations of hazards for a province~\cite{ullah2022multi}.
\item \textit{Single-model-from-ensemble strategy}: Random forest was found to be best performing amongst a set of ML models (boosted regression tree (BRT), RF, and support vector machine (SVM)) for susceptibility maps of flooding, gully erosion, and forest fires~\cite{pouyan2021multi}. The three maps along with the PGA map for earthquakes were reduced from multi-class severity to 2-class and then combined using different combinations of the presence or absence of each hazard. For a state-level MHSM, RF and SVM are used for computing LSM and FSM separately~\cite{nachappa2020multi}. The MHSM based on RF and SVM are computed separately using a bivariate color scheme and compared.
\item \textit{Optimal-models strategy}: Logistic regression or random forest is used for LSM computation, modified-AHP is used for FSM, and PGA is computed for earthquake hazard map, and finally the maps are combined using Mamdani fuzzy algorithm~\cite{karakas2023hybrid,yanar2020use} for cases of urban settlements and basins. For a study in a mountainous region, the generalized linear model (GLM) is used with logistic regression for LSM computation, AHP is used for FSM computation, and PGA map for earthquakes, which are combined using weighted sum, with weights obtained from AHP~\cite{aksha2020geospatial}. For an MHSM of a mountainous region, SVM is used for susceptibility maps of landslides, land subsidence, and floods, GLM for wildfires, and functional discriminant analysis (FDA) for snow avalanches~\cite{yousefi2020machine}. The different maps are then combined using a boolean algorithm for computing MHSM.
\end{itemize}

\subsection{A Generalized Algorithm}
We can now summarize the MHSM computation based on a late fusion strategy in the form of a high-level generalized algorithm (Algorithm~\ref{alg:mhsm}). The algorithm is oversimplified here to demonstrate the late fusion strategy in the state-of-the-art methods.

\begin{algorithm}[t]
  \small
  \caption{\label{alg:mhsm} A High-level Generalized Algorithm for Multi-Hazard Susceptibility Mapping}
  \DontPrintSemicolon
  \SetAlgoLined
  \SetKwInOut{Input}{Input}
  \SetKwInOut{Output}{Output}
  \Input{Heterogeneous data sources for conditioning factors for $N$ natural hazards}
  \Input{Inventory for $n$ natural hazards ($n\le N$)}
  \Output{A raster image of the map of the study area, $O_h$, colored based on the choice of combining multiple single-hazard susceptibility mapping}
  \For {hazard $0 <  h \le N$}
       {
         Create single-hazard susceptibility map $O_r^h$\;
       } \;
  Create $O_h$ by combining \{$O_r^1$, $O_r^2$, $\ldots$, $O_r^N$\} single-hazard susceptibility maps using weighted averaging, simple averaging, (thresholding and addition), or bivariate color scheme \;
  \KwRet $O_h$\;
 \end{algorithm}
\section{Background on Deep Learning Models for Susceptibility Mapping}\label{sec:dl}

Deep learning models are emerging as effective solutions in single-hazard susceptibility mapping, \eg for floods, landslides, etc.~\cite{bentivoglio2022deep,fang2021comparative,costache2020novel,pham2021can}, and in multi-hazard susceptibility mapping~\cite{ullah2022multi}.

\subsection{Single-Hazard Susceptibility Mapping}
Increasingly deep neural networks (DNNs), such as multilayer perceptrons (MLPs) and CNNs, are widely used for FSM~\cite{bentivoglio2022deep}. MLP networks are neural networks composed of fully connected layers, which use a one-directional layer propagation rule based on a point-wise nonlinear function, thus providing a \textit{feed-forward} mechanism. These nonlinear functions include sigmoid, rectified linear unit (ReLU), max function, etc. MLPs also use a \textit{backpropagation} mechanism to fine-tune the weights by propagating errors from the output into the neural network. 

Inductive biases allow the reuse of parameters in different parts of the input of each layer. For susceptibility mapping, spatial equivariance is an inductive bias that can be exploited, as the neural network improves the outcomes based on spatial continuity. Temporal equivariance is another inductive bias that can be exploited for spatiotemporal datasets, however, the current methodology for susceptibility mapping does not capture temporal patterns. Fully connected layers, and hence MLPs, cannot have any inductive bias capabilities. Convolutional kernels work with images of different dimensions and recurrent layers work with time series of variable lengths. Hence, the spatial and temporal inductive biases apply to CNNs and recurrent neural networks (RNNs), respectively.

Owing to the spatial equivariance as an inductive bias, CNNs tend to be effective for FSM, especially when raster images are used as inputs (Algorithm~\ref{alg:shsm}). Convolution is a signal processing operation where an entry is replaced by the weighted average of its neighbors, where the weights come from a kernel/signal. Convolution, when applied to images, gives spatially weighted average values of neighbors of a pixel for itself. The spatial equivariance inductive bias helps in the reuse of kernels across different parts of the input image. Given the design of processing sequential data, \eg time series, RNNs are not used for susceptibility mapping despite their inductive bias. MLPs are effectively used for susceptibility mapping by coupling with SA models to address the absence of inductive bias.

\begin{figure}
\centering
\includegraphics[width=\linewidth]{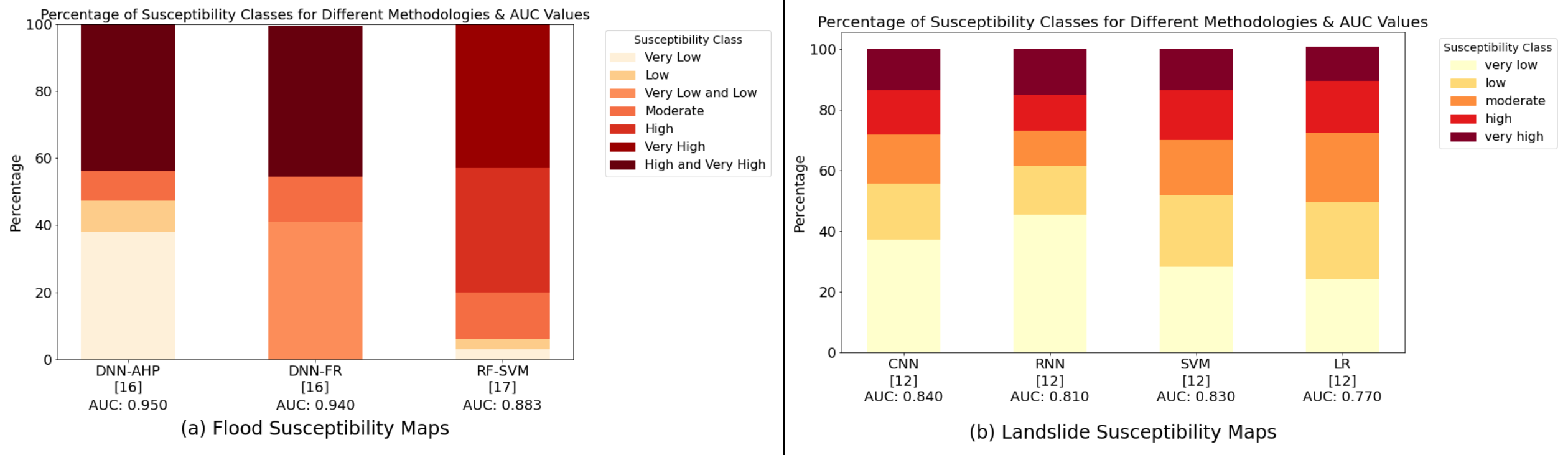}
\caption{\label{fig:shsm-results} Comparative results of ML/DL models used for single-hazard susceptibility mapping in terms of the percentage of area in different levels of severity of the natural hazard and the area under the ROC curve (AUC) metric for (a) FSMs~\cite{costache2020novel,nachappa2020flood} and (b) LSMs~\cite{fang2021comparative}.}

\includegraphics[width=\linewidth]{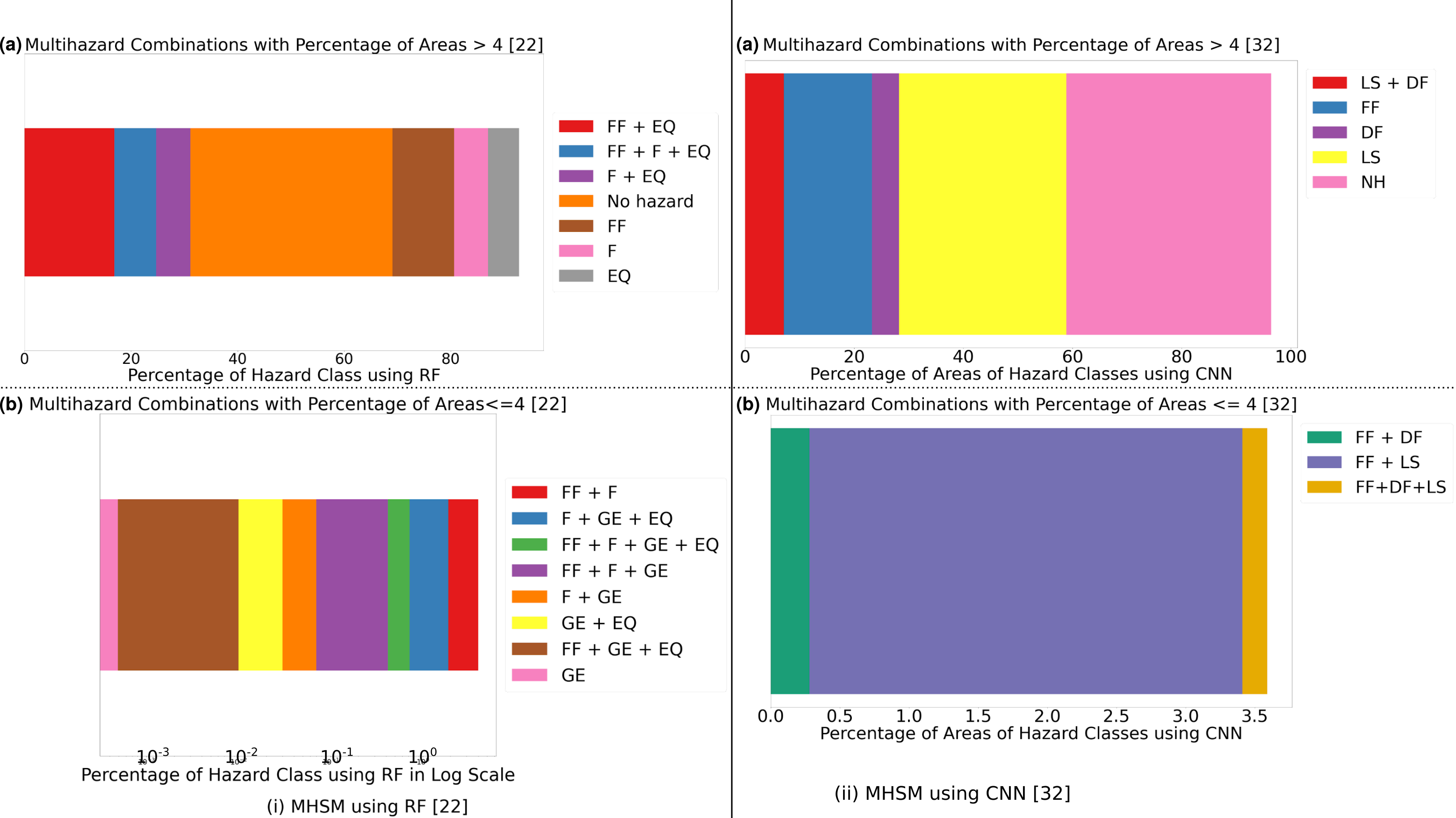}
\caption{\label{fig:mhsm-results} Comparative results of ML/DL models used for multi-hazard susceptibility mapping in terms of percentage of combinations of the natural hazards for (i)  ML (using RF)~\cite{pouyan2021multi} for flash floods, earthquakes, floods, and gully erosion, and (ii) DL (using CNNs)~\cite{ullah2022multi} for landslides, debris flow, and flash flood. The charts are in linear scale for percentages (a) $>4$, and (b) $\le 4$, except for the logarithmic scale in (ii),(b). NH implies `no hazard.}
\end{figure}

DL models can be used in an ensemble where multiple models are combined to improve the outcomes. For instance, a DNN can be combined with an MCDA model (\eg AHP) or SA model (\eg FR) by using AHP weights or FR coefficients in the DNN for flash flood susceptibility mapping~\cite{costache2020novel}. Ensemble learning can also be incorporated amongst ML/DL models, which include methodologies such as stacking, blending, simple averaging, and weighted averaging~\cite{fang2021comparative} of the \textit{base classifiers}, \ie the models in the ensemble. Stacking involves re-organizing the outcomes of the base classifiers from K-fold cross-validation as new features, which form the new meta-training data, and running a meta-classifier. A blending ensemble is a variant of stacking where the base classifiers are trained and the outcomes are concatenated to form the new meta-training data to be used in the meta-classifier. In both stacking and blending, the outcomes of the meta-classifier are final. In simple averaging, the output probabilities from the base classifiers are averaged, and in weighted averaging, the AUC values of the corresponding models are used as weights for averaging the output probabilities of the base classifiers. For a case study, weighted averaging outperformed other ensemble learning methods for an LSM computation~\cite{fang2021comparative}.

One of the issues with DL models is the lack of studies on their generalizability for susceptibility mapping. If DL models can be pretrained for a region and transferred to new areas, then such models are economical. However, if DL models are not generalizable, then it is cost-intensive to train new models for new areas. Another issue with DL models is about being data-hungry and resource-intensive. Data-driven susceptibility mapping methods address this issue by using heterogeneous data sources. Running these models on the graphics processing units (GPUs) makes them relatively more efficient~\cite{bentivoglio2022deep}. A critical issue with the use of deep learning models for susceptibility mapping is the lack of standardized guidelines for the choice of conditioning factors and appropriate model selection~\cite{pham2021can}.

We compare results from case studies in FSMs~\cite{costache2020novel,nachappa2020flood} and  LSMs~\cite{fang2021comparative} in Figure~\ref{fig:shsm-results}, where we juxtaposed results from ML/DL models. We observe that DL models consistently have higher AUC values for FSM and LSM. For LSM, SVM also is comparable to DL in one such case study~\cite{fang2021comparative}.

\subsection{Multi-hazard Susceptibility Mapping}

For data-driven MHSM computation, there is only one study that uses DL through the late fusion of CNNs~\cite{ullah2022multi}. Its results are compared to those of an ML model~\cite{pouyan2021multi} for MHSM in Figure~\ref{fig:mhsm-results}. Qualitatively, they are comparable results since the results are presented using visualizations.

\section{Vision for DL in MHSM and Discussion} \label{sec:newvision}

In the context of the emergence of DL methods for single-hazard susceptibility mapping, we propose the vision of adopting several data fusion strategies used for multimodal data in their DL models~\cite{zhang2021deep}. Our rationale behind the vision is to expand the design space of DL methods in multi-hazard susceptibility mapping to improve its adoption.

Since the outcome is a colored raster image of a geospatial map, we pose the single- and multi-hazard susceptibility mapping as a \textit{semantic segmentation} of images, which can also be referred to as \textit{pixel-level classification}. For MHSM computation, we learn from multiple single-hazard susceptibility maps, which can be treated as multimodal input to the semantic segmentation problem. In computer vision, there are advancements in several fusion strategies for deep multimodal learning.

\begin{figure}[t]
  \centering
  \includegraphics[width=\textwidth]{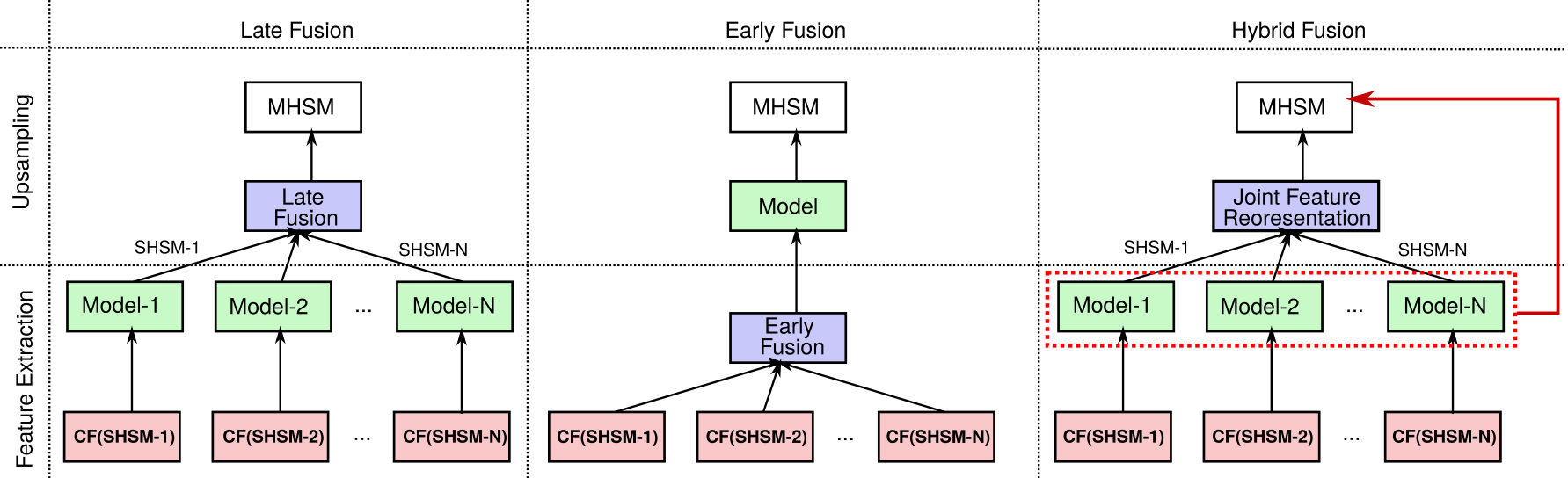}
  \caption{\label{fig:mhsm-dl} MHSM computation using different fusion strategies in deep multimodal learning when using conditioning factors (CFs) of multiple single-hazard susceptibility maps (SHSMs). Here, \textbf{CF(SHSM-i)} refers to a set of raster images of CFs for the $i$\textsuperscript{th} SHSM. (Adapted from~\cite{zhang2021deep})}
\end{figure}

The three classic strategies for fusion are late, early, and hybrid fusion (Figure~\ref{fig:mhsm-dl}). Late fusion involves combining decisions, as is the case of the state-of-the-art MHSM computation using DL. Early fusion involves data- or feature-level fusion, and hybrid fusion methods optimize the strengths of early and late fusion. An early fusion strategy can reduce the cost of learning by utilizing a single segmentation network. In comparison, the late fusion strategy is convenient as the segmentation for each hazard is de-coupled and can reuse the well-researched stable process followed in the domain. The hybrid fusion strategy allows the segmentation network to access data through its branch. This introduces an extra module for class-wise or modality-wise (\ie hazard-wise) weights, thus leading to a joint feature representation.

In the context of MHSM computation, designing late, early, and hybrid strategies are in the order of increasing complexity. We hypothesize that the early data fusion strategy is a worthwhile strategy to exploit inter-relationships in conditioning factors. The strategy is implementable given the uniformity in raster image inputs, especially when they are accessible through integrator tools, such as Google Earth Engine, and ESRI ArcGIS. Early fusion helps in directly exploiting inter-relationships at the data level as not all the conditioning factors are independent variables.

One of the limitations of our proposed method is that the three strategies yield different results, and hence, they must be used based on the semantics of the problem statement. When we perform early fusion at the data level, the outcome of MHSM is metric-based, where a gradient color palette is then used for visualization~\cite{rehman2022multi,karakas2023hybrid}. As we observe in the late fusion, multiple single-hazard susceptibility maps preserve themselves. Late fusion results are visualized using a categorical color palette, which is based on combinations of the contributing hazards. Feature fusion leads to a combination of a few multiple hazards, which is the same as late fusion. However, feature fusion of other hazards is similar to early fusion. Thus, depending on the process of MHSM computation, the outcomes of the feature fusion may vary. Overall, it may be difficult to compare the results of the three strategies of multimodal data fusion. A workaround is to use a \textit{standardized} multi-hazard susceptibility metric across all strategies.

\section{Concluding Remarks}\label{sec:conclusions}

In this work, we have provided a brief overview of data-driven methods for single- and multi-hazard susceptibility mapping. Our in-depth study provides high-level generalized algorithms for both. We then analyzed the use of deep learning in both and found that research in single-hazard susceptibility mapping actively uses deep learning (DL). To improve the adoption of deep learning in multi-hazard susceptibility mapping, we posed DL-based multi-hazard susceptibility mapping to be akin to deep multimodal learning for semantic segmentation of images. We then proposed the vision of the use of different fusion strategies for DL-based multi-hazard susceptibility mapping to expand its scope. This work only scratches the surface of fusion strategy in multi-hazard susceptibility mapping which needs to be studied in the future to understand its value.

\bibliographystyle{unsrt}
\bibliography{papers_mhm,papers_flood}
\end{document}